# КОМПЛЕКСНЫЕ ИНСТРУМЕНТАЛЬНЫЕ СРЕДСТВА ИНЖЕНЕРИИ ОНТОЛОГИЙ

**Виталий Величко, Кирилл Малахов, Виталий Семенков, Александр Стрижак**

*Аннотация: В статье представлен обзор актуальных специализированных инструментальных средств инженерии онтологий, а также средств аннотирования текстов на основе онтологий. Рассмотрены основные функции и возможности данных инструментальных средств, их достоинства и недостатки. Описываются основные компоненты системы формирования онтологий на основе семантического анализа текстовых массивов. Дается системный сравнительный анализ средств инженерии онтологий.*

## Введение

Методология проектирования онтологии предметной области (ПрО) [Палагин, 2012] основывается на конструктивном использовании следующих категорий – множества концептов, отношений, функций интерпретации и аксиом. Построение указанных множеств является трудоёмким процессом, как по времени, так и по количеству вовлечённых в процесс проектирования высококвалифицированных специалистов. Особенно трудоемко ручное проектирование онтологий, которое мало чем отличается от тематического проектирования экспертных систем.

Понимание важности решения проблемы создания эффективных инструментальных средств поддержки процессов проектирования онтологии заданной ПрО, пришло практически одновременно с осознанием парадигмы компьютерных онтологий. В настоящее время известно более ста инструментальных программных систем [Гаврилова, 2000; Палагин, 2009], но только ограниченное количество комплексных программных систем включают редактор онтологических структур, автоматизированное построение онтологий ПрО, средства поверхностного семантического анализа текстовых документов, используемых для построения онтологии.

## Основные характеристики комплексных программных систем проектирования онтологий

К основным характеристикам комплексных программных систем онтологического инжиниринга можно отнести: поддерживаемые формализмы и форматы представления онтологий, архитектура программного обеспечения, интерфейс пользователя, функционал редактора онтологии, средства хранения онтологий, доступность, дополнительные возможности. Ниже рассмотрим более подробно некоторые из приведенных характеристик.

### Поддерживаемые формализмы и форматы представления онтологий

Под формализмом понимается некоторая формальная теория [Клини, 1957], лежащая в основе способа представления онтологических знаний (логика предикатов, фреймовые модели, дескриптивная логика, концептуальные графы и др.). Выбранный формализм существенно влияет на организацию внутренних (компьютерных) структур данных и может определять их формат представления.

Формат представления онтологий задаёт вид их хранения в библиотеке, способ передачи онтологических описаний другим потребителям и метод обработки ее концептов. В качестве форматов онтологических описаний разработаны определенные языки представления онтологий, наиболее известными из которых являются OWL, RDFS, KIF.

Некоторые из известных редакторов онтологий поддерживают работу с несколькими формализмами представления, однако следует учитывать тот факт, что обычно конкретный формализм является предпочтительным для конкретного редактора [Палагин, 2009].

### Функциональность редактора онтологии

Функциональность редактора онтологии является одной из самых важных характеристик, под которой понимается множество предоставляемых пользователю сервисов работы с онтологическими структурами.

Базовый набор функций редактора онтологии обычно обеспечивает:

- работу с одним или несколькими онтологическими описаниями (проектами) одновременно;
- графический интерфейс пользователя;
- редактирование онтологии (создание, редактирование, удаление концептов, отношений, аксиом и прочих структурных элементов онтологии);
- инкапсулирование онтологий в среду информационных систем.

### Дополнительные возможности

К дополнительным возможностям относят поддержку языка запросов, анализ целостности, использование механизма логического вывода, поддержку удалённого доступа через Интернет, документирование.

Известны три группы инструментальных средств (ИнС) онтологического инжиниринга [Овдей, 2006]. К первой группе относят инструменты создания онтологий, которые предполагают поддержку совместной разработки и просмотра, создание онтологии в соответствии с заданной (произвольной) методологией, поддержку рассуждений.

Ко второй группе относят инструменты объединения, отображения и выравнивания онтологий. Объединение предполагает нахождение сходств и различий между исходными онтологиями и создание результирующей онтологии, которая содержит элементы исходных онтологий. Для этого ИнС автоматически определяют соответствия между концептами или обеспечивают графическую среду, в которой пользователь сам находит эти соответствия. Процедура отображения заключается в нахождении семантических связей между концептами различных онтологий. Процедура выравнивания онтологий устанавливает различные виды соответствия между двумя онтологиями, информация о которых сохраняется для дальнейшего использования в приложениях пользователя [Noy, 1999].

К третьей группе относят инструменты для аннотирования Web-ресурсов на основе онтологий.

Содержательный обзор известных инструментов инженерии онтологий, в котором рассмотрены основные функции и возможности ИнС, их достоинства, недостатки, сравнительный анализ и описание известных доступных онторедакторов, также приведен в [Овдей, 2006; Noy, 1999; Calvanese, 2007; Филатов, 2007].

Общими недостатками большинства известных инструментальных средств являются:

- отсутствие процедур автоматического (автоматизированного) формирования компонент онтологии;
- англоязычный интерфейс с пользователем, в котором (для большинства ИнС) не предусмотрено присвоение имён компонентам онтологии на русском или украинском языке;
- структуризация концептов выполняется только по одному типу отношений;
- для большинства общедоступных ИнС не предусмотрена работа с большими по объёму онтологиями (например, для OntoEditFree – до 50 концептов);
- большинство инструментов хранит свои онтологии в текстовых файлах, что ограничивает скорость доступа к онтологиям;
- задекларированные функциональные возможности для общедоступных инструментов зачастую так и остаются нереализованными;
- недостаток информации для пользователей в инструкциях.

Рассмотрим более подробно некоторые инструментальные программные системы, предназначенные для построения онтологий и их использования для решения задач. Универсальные и специализированные оболочки программных систем являются средством, упрощающим процесс создания интеллектуальной системы [Артемьева, 2008]. Универсальные оболочки основаны на использовании некоторого универсального языка представления знаний. В специализированных оболочках при представлении знаний используется специфичная для предметной области схема, определяемая онтологией той области, для которой создается оболочка, что позволяет создавать базу знаний эксперту предметной области без участия посредника, которым является инженер знаний. В сложно-структурированных предметных областях, связанных с наукой, могут изменяться не только знания, но и онтологии, и, как следствие, множество классов решаемых задач. Описание особенностей специализированных оболочек интеллектуальных систем для сложно-структурированных предметных областей приведено на основе статьи «Интеллектуальная система, основанная на многоуровневой онтологии химии» [Артемьева, 2008].

## Специализированная оболочка интеллектуальной системы для сложно-структурированных предметных областей

Информационными компонентами специализированной оболочки для сложно-структурированной ПрО являются многоуровневая модульная онтология и модульная база знаний. Создание и редактирование информационных компонент осуществляется многоуровневым редактором онтологий и редактором знаний, разработка которых основывается на онтологии уровня *n*.

Редакторы многоуровневых онтологий и знаний должны позволять создание и редактирование модульных онтологий и знаний, а также обеспечивать возможность повторного использования модулей при создании онтологий и знаний новых разделов и подразделов области, причем процесс создания и редактирования модуля онтологии уровня *i*-1 должен управляться онтологией уровня *i*, а процесс создания и редактирования модуля знаний – онтологией уровня 2.

Редактор онтологии должен обеспечивать возможность выбора того из существующих модулей онтологии уровня *i*, который управляет процессом редактирования создаваемого модуля. Аналогично при редактировании модуля знаний должна обеспечиваться возможность выбора «управляющего» модуля онтологии уровня 2.

Редакторы онтологии и знаний должны обеспечивать возможность задания структурированной и неструктурированной части онтологии, а также структурированной и неструктурированной части знаний, т. е. программным компонентом этих редакторов должен быть специализированный редактор утверждений, позволяющий вводить онтологические соглашения и законы предметной области.

Редактор знаний должен обеспечивать возможность ввода/вывода значений нестандартных величин при редактировании знаний. Для значений нестандартных величин в предметной области может существовать способ их графического представления. Например, для химии [Артемьева, 2008] графически может быть задана краткая структурная формула или структурная формула химического соединения. Поэтому редактор знаний должен обеспечивать возможность использования принятого в предметной области графического способа представления значений нестандартных величин при создании и редактировании знаний. Величина, которой принадлежит значение некоторого свойства, задается онтологией уровня 2. Поэтому редактор знаний должен обеспечивать автоматический выбор (управляемый онтологией уровня 2) средств для графического представления значений нестандартных величин при редактировании знаний.

Редактор онтологии интерпретирует онтологию уровня *i* при создании модуля онтологии уровня *i*-1. Редактор знаний интерпретирует онтологию уровня 2 при создании модуля знаний. Одна и та же онтология может интерпретироваться разными способами в разных редакторах знаний. Редакторы знаний могут отличаться не только способом интерпретации знаний, но и интерфейсом. Очевидно, что более удобный интерфейс и более понятный эксперту способ интерпретации можно обеспечить для редактора, предназначенного для интерпретации одной онтологии, а не класса онтологий. Поэтому специализированная оболочка должна позволять использование редакторов, поддерживающих разные способы интерпретации модуля онтологии уровня 2 и предоставлять возможность эксперту выбора требуемого ему редактора знаний.

Значения нестандартных величин используются не только при редактировании знаний, но также при вводе исходных данных задач. Графический способ задания исходных данных задач более удобен для специалиста предметной области, поскольку в этом случае отсутствует необходимость громоздкого вербального описания этих данных. Графическое представление результатов решения является более наглядным способом представления. Поэтому оболочка должна обеспечивать возможность ввода/вывода значений нестандартных величин при задании исходных данных задач, а также позволять использование принятого в предметной области графического способа представления значений нестандартных величин при вводе исходных данных задач и выводе результатов их решения.

Как уже отмечалось, величина, которой принадлежит значение некоторого свойства, задается онтологией уровня 2. Оболочка должна обеспечивать автоматический выбор (управляемый онтологией) средств для графического представления значений нестандартных величин при задании исходных данных задач.

Каждый раздел сложно-структурированной ПрО характеризуется своим множеством классов прикладных задач, причем разные множества могут содержать как общие классы задач, так и специфичные для раздела. Решатель задач может быть предназначен для решения классов задач одного раздела (в этом случае он использует онтологию и знания этого раздела), либо разных разделов (в этом случае он может использовать разные онтологии и знания). В первом случае используемая решателем онтология определяется классом задач. Во втором случае требуется дополнительное указание, какие онтология и знания должны использоваться в процессе решения. Специализированная оболочка интеллектуальных систем для сложно-структурированной предметной области должна обеспечивать возможность решения задач разных классов, причем пользователь должен иметь возможность указания модуля онтологии и модуля знаний, которые надо использовать при решении задач.

Таким образом, специализированная оболочка должна содержать расширяемые библиотеки систем для решения задач разных классов, системы автоматического построения методов решения задач по их спецификации (рис. 1).

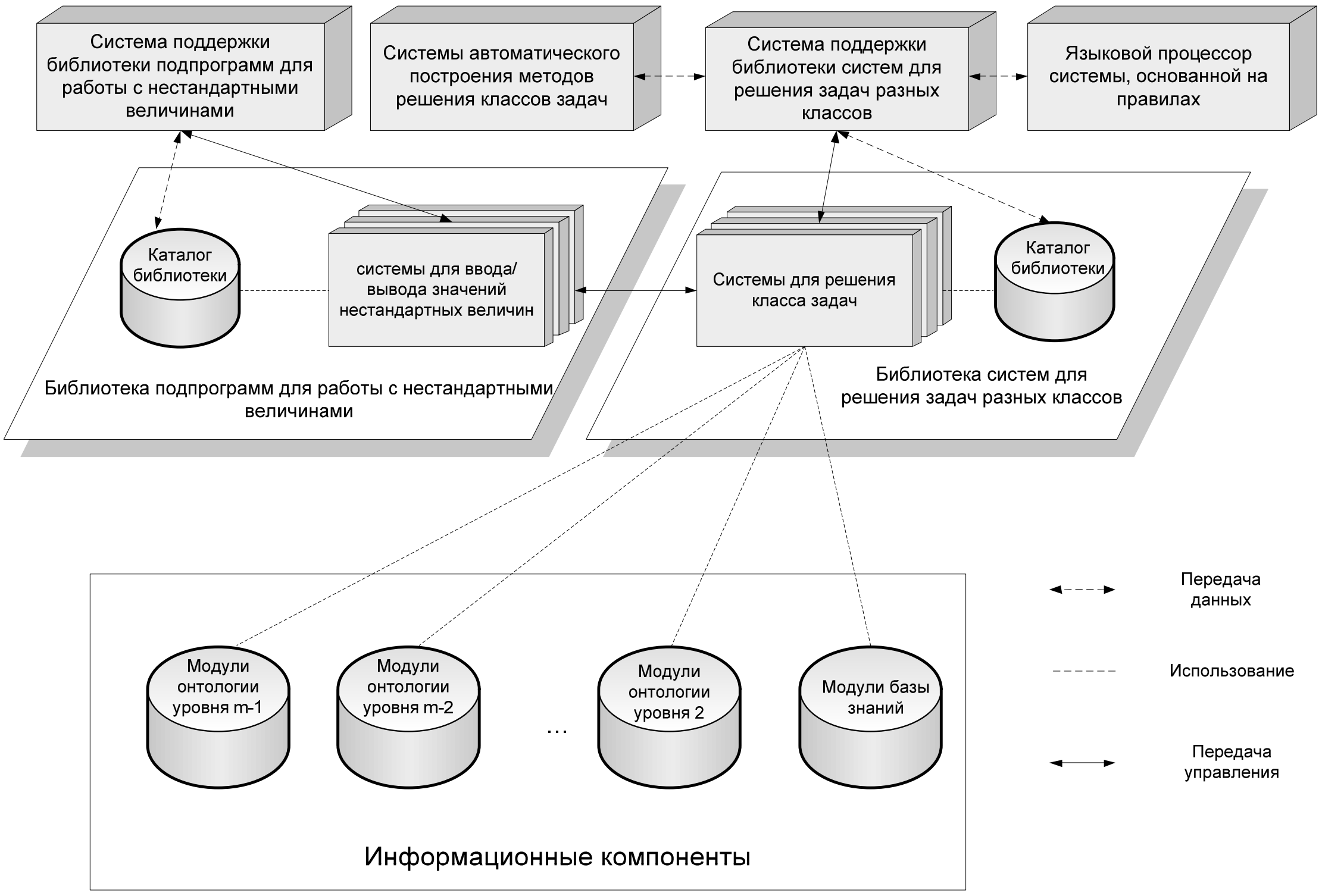

Рис. 1. Системы для решения задач и средства их разработки

Метод решения задач может быть представлен либо в виде алгоритма, либо в виде множества правил системы продукций. В первом случае для создания решателя задач используется процессор алгоритмического языка, во втором случае – процессор языка, основанного на правилах, который является одним из программных компонент специализированной оболочки.

## Специализированная система обработки текстовых документов «ЛоТА»

Специализированная система обработки текстовых документов «ЛоТА» [Невзорова, 2001] является системой класса Text Mining. Система предназначена для анализа специализированных текстов «Логика работы», описывающих логику работы сложной технической системы в различных режимах функционирования. Основной задачей анализа является извлечение из данных текстов информационной модели алгоритмов, решающих определенную задачу в определенной проблемной ситуации, и контроль структурной и информационной целостности выделенной схемы алгоритмов.

Информационная модель алгоритма включает:

- описание входного информационного потока (типы информационных сигналов или семантическое описание информационного потока с указанием источника информации – конкретный алгоритм, конкретное измерительное устройство);
- описание процессов преобразования входных данных в выходные (допустимый способ разрешения проблемы);
- описание выходного информационного потока (типы информационных сигналов или семантическое описание информационного потока с указанием точки приема информации).

Решение основной задачи обеспечивается комплексом технологий обработки текстов, включающих:

- технологии морфосинтаксического анализа;
- технологии семантико-синтаксического анализа;
- технологии взаимодействия с прикладной онтологией.

Указанная сумма технологий формируется на основе центрального ядра – прикладной онтологии (в дальнейшем, авиаонтология [Лукашевич, 2004]), обеспечивающей согласованное взаимодействие различных программных модулей. Авиаонтология концептуально описывает предметную область информационного обеспечения различных полетных режимов антропоцентрических систем. Авиаонтология представляет собой сеть понятий предметной области. Авиаонтология относится к классу лингвистических (лексических) онтологий и предназначена для встраивания в различные лингвистические приложения.

Программный комплекс состоит из трех взаимодействующих подсистем: подсистемы лингвистического анализа технических текстов «Анализатор», подсистемы ведения онтологии «OntoEditor+» и подсистемы «Интегратор». Взаимодействие подсистем реализовано на базе технологии «клиент-сервер», причем в различных подзадачах подсистемы выступают в различных режимах (режим сервера или режим клиента).

Инструментальная система визуального проектирования «OntoEditor+» является специализированной СУБД. Система предназначена для ручного редактирования онтологий, хранящихся в реляционной базе данных в формате TPS, а также обслуживания запросов пользователей и внешних программ. Новые возможности системы обеспечиваются функциональным набором «Лингвистический инструментарий», посредством которого реализуется встраивание прикладной онтологии в лингвистические приложения. Наиболее типичными задачами, решаемыми с помощью инструментария системы «OntoEditor+», являются: изучение структурных свойств прикладной онтологии с помощью исследовательского инструментария системы «OntoEditor+»; построение лингвистической оболочки прикладной онтологии; задача покрытия текста онтологическими входами; построение выводов по прикладной онтологии и др.

Подсистема «Анализатор» реализует основные этапы лингвистической обработки текста (графематический, морфосинтаксический и частичный синтаксический анализ).

Подсистема «Интегратор» исполняет внешний запрос на извлечение знаний из текста. Структура внешнего запроса содержит компоненты информационной модели алгоритма. Внешний запрос интерпретируется при взаимодействии с подсистемой «OntoEditor+» как структура, привязанная к прикладной онтологии. Выделение компонент информационной модели происходит на основе механизмов отождествления элементов дерева сегментов входного текста (взаимодействие с подсистемой «Анализатор») и элементов структуры запроса (взаимодействие с подсистемой «OntoEditor+»).

Инструментальная система визуального проектирования онтологий «OntoEditor+» включает:

- лингвистический инструментарий (задачи корпусного исследования (загрузка корпуса; сегментация на предложения; автоматическое ведение статистики по различным объектам корпуса), построение лингвистической оболочки онтологии, задача покрытия текста онтологическими входами, построение выводов по онтологии, поддержка протоколов информационного обмена системы «OntoEditor+» с внешними программными модулями, в том числе с внешними информационными ресурсами);
- исследовательский инструментарий.

Основные функции подсистемы «Анализатор»:

- графематический анализ;
- морфосинтаксический анализ;
- покрытие текста онтологическими входами (взаимодействие с системой «OntoEditor+»).

Основные функции подсистемы «Интегратор»:

- анализ и исполнение внешнего запроса (информационная модель алгоритма);
- интерпретация внешнего запроса в терминах прикладной онтологии (взаимодействие с системой «OntoEditor+»);
- интерпретация внешнего запроса в структурных компонентах дерева сегментов (взаимодействие с системой «Анализатор»);
- контроль информационной целостности (анализ компонент внешнего запроса).

Метод контекстного разрешения омонимии является базовым методом в интегральной технологии разрешения омонимии в системе «ЛоТА». Однако практические задачи системы выявили ряд важных аспектов лингвистического анализа, которые стимулировали развитие новых методов разрешения многозначности.

Интегральная технология разрешения многозначности, разрабатываемая в системе «ЛоТА», включает следующие методы:

- метод контекстного разрешения функциональной омонимии;
- метод разрешения функциональной, грамматической и лексической омонимии на основе индексируемой базы устойчивых коллокаций;
- метод разрешения функциональной, грамматической и лексической омонимии на основе лингвистической оболочки онтологии.

Для эффективного встраивания в лингвистические приложения система «OntoEditor+» поддерживает группу протоколов информационного обмена с внешними программными модулями системы и внешними словарными базами данных, обеспечивая работу в режиме клиент-сервер. Разрешение многозначности (функциональной, морфологической и лексической) во входных текстах происходит на основе механизма распознавания контекстов омонимов, зафиксированных в индексируемой базе контекстов.

Разработаны три основных механизма пополнения индексируемой базы контекстов функциональных омонимов:

- ручной ввод и редактирование данных по типовым контекстам омонимов;
- импорт типовых контекстов омонимов из текстового файла, подготовленного в специальном формате представления данных;
- импорт типовых контекстов омонимов, обнаруженных специальными механизмами поиска подсистемы «Анализатор».

Данный механизм организован как запрос к подсистеме «Анализатор» с передачей ему от подсистемы «OntoEditor+» текстового корпуса, по которому проводится поиск. В процессе обработки подсистема «Анализатор» передает подсистеме «OntoEditor+» информацию об обнаруженных контекстах омонимов, которая записывается либо в индекс омонимов, либо в автоматическом режиме, либо в режиме диалога с оператором. Отличительной особенностью режима диалога является режим самообучения, который реализуется с использованием механизма журнала событий. В данном журнале в зависимости от его настройки фиксируются те или иные важные события в системе, например, изменение информации в индексе омонимов или операции взаимодействия с подсистемой «Анализатор». В режиме самообучения сохраняется и контролируется последовательность ранее сгенерированных диалогов, что обеспечивает генерацию только уникальных диалогов на разрешение омонимии без повторений.

Лингвистический инструментарий подсистемы «OntoEditor+» обеспечивает встраивание онтологии в различные приложения, связанные с обработкой текстов. Лингвистический инструментарий реализует функции загрузки корпуса текстов; автоматическое ведение статистики по различным объектам корпуса; функции предсинтаксической обработки текста (сегментация предложений, распознавание аббревиатур, разрешение омонимии на основе специальных протоколов взаимодействия с внешними словарными ресурсами); построение лингвистической оболочки онтологии; распознавание терминов прикладной онтологии во входном тексте (задача покрытия). Сопряжение онтологического и лингвистического (грамматического) ресурсов реализуется через механизмы лингвистической оболочки онтологии. Лингвистическая оболочка онтологии создается с помощью разработанного программного инструментария, посредством которого фиксируется грамматическая информация об онтологических концептах и их текстовых формах. Каждый онтологический вход (как правило, многословный термин) снабжается соответствующей грамматической информацией, при этом для омонима разрешается соответствующая (функциональная, лексическая, морфологическая) омонимия. Грамматическая информация передается в подсистему «OntoEditor+» от подсистемы «Анализатор» на основе специальных протоколов взаимодействия. Разрешение лексической, функциональной и морфологической омонимии выполняется на основе специальных диалогов с экспертом-лингвистом. Отдельные процедуры реализуют проверки словоформ в составе терминологического входа на согласованность их грамматических характеристик, также осуществляется контроль достоверности словарной информации. Контроль достоверности обеспечивает отслеживание изменений, как в составе грамматического словаря, так и в составе онтологии. Учитывая сложность и многоступенчатость вышеперечисленных процедур, в подсистеме «OntoEditor+» разработан мастер построения лингвистической оболочки, который вызывается командой основного меню.

Подсистема «Анализатор» обеспечивает реализацию метода разрешения омонимии на основе контекстных правил, т. е. фактически используются лингвистические знания системы. Этот метод является универсальным, не зависит от специфики предметной области и обеспечивает в текущей версии точность распознавания не ниже 95 %. Однако, для данного метода существуют крайне сложные типы функциональной омонимии, например, тип «частица/союз». Разрешение данной омонимии возможно во многих случаях лишь после завершения полного синтаксического анализа.

Взаимодействие подсистемы «OntoEditor+» и подсистемы «Анализатор» осуществляется на основе специальных протоколов взаимодействия. При применении интегральной технологии разрешение многозначности происходит в два этапа. На первом этапе подсистема «Анализатор» (клиент) передает запрос на разрешение омонимии входного текста подсистеме «OntoEditor+» (сервер). Подсистема «OntoEditor+» возвращает подсистеме «Анализатор» информацию о разрешенных омонимах на основе своих методов. На втором этапе подсистема "Анализатор" разрешает омонимию оставшихся неразрешенных омонимов на основе метода контекстных правил.

Интегральная технология разрешения многозначности эффективно применяется на этапе предсинтаксического анализа в системе «ЛоТА». По существу, интегральная технология представляет собой сочетание инженерного и лингвистического подхода к решению поставленной задачи. В основе проектирования интегральной технологии лежат процессы скоординированного взаимодействия различных языковых уровней, прежде всего онтологического уровня (обеспечивающего системные модели знаний о мире) и различных языковых уровней (морфологического и синтаксического). В системе реализован эффективный механизм взаимодействия различных подсистем, обеспечивающих реализацию различных методов в составе интегральной технологии. При этом, следует признать, что сам процесс согласования языковых взаимодействий достаточно сложен и требует определения большого числа дефиниций-правил, описывающих условия обработки концептов онтологии.

## Интеллектуальная система извлечения данных и их анализа (на основе текстов) ИСИДА-Т

Целью ИСИДА-Т [Кормалев, 2006]; [Киселев, 2004], является извлечение значимой информации определенного типа из (больших массивов) текста для дальнейшей аналитической обработки. Результатом работы систем является получение структурированных данных и отношений на них.

Основные компоненты ИСИДА-Т:

- Инфраструктурные службы (конфигурирование, параллельная обработка, взаимодействие модулей);
- Лингвистический процессор;
- Модули работы со знаниями ПрО;
- Интерпретатор правил извлечения информации.

Разработанные в рамках проекта ИСИДА-Т технологии, инструменты и продукты позволяют:

- обнаруживать в электронных документах, извлекать и структурировать информацию о представляющих интерес фактах, событиях, объектах и отношениях;
- выполнять мониторинг сайтов в сети Интернет на предмет появления там значимой для пользователя информации.

Основные рабочие характеристики технологии и продуктов:

- поддержка русского языка;
- быстрая настройка на предметную область при помощи эффективных инструментальных средств;
- высокая точность и полнота анализа за счет использования предметных знаний;
- наличие встроенных средств визуализации результатов анализа в виде диаграмм и схем;
- легкая интегрируемость в другие информационные системы на любом уровне (программный или сетевой интерфейс, БД);
- функционирование под управлением ОС Windows и большинства Linux-систем;
- близкая к линейной масштабируемость при параллельной архитектуре анализа. Возможность работы на вычислительных машинах кластерного типа.

Некоторые области применения технологий семантического анализа и структурирования текстовой информации:

- информационная поддержка бизнеса (business intelligence) и управление знаниями (knowledge management);
- маркетинговые исследования;
- финансовая аналитика;
- военная и коммерческая разведка и мониторинг;
- информационная поддержка органов государственной власти (в рамках направления «Электронное правительство»);

- работа библиотек, издательств и СМИ.

Рассмотрим общую организацию инфраструктуры системы ИСИДА-Т. Краеугольным камнем системы ИСИДА-Т является точная настройка на предметную область и конкретную задачу извлечения. С одной стороны, это достигается за счет редактирования лингвистических ресурсов, ресурсов знаний, правил извлечения и правил трансформации. С другой стороны, настройка может потребовать включения в процесс обработки дополнительных специализированных методов обработки текста. Кроме того, для каждой задачи необходимо подобрать наиболее подходящие алгоритмические средства анализа из набора имеющихся. Эти аспекты требуют создания такой архитектуры, при которой легко могут добавляться и замещаться алгоритмические компоненты процесса извлечения.

Проблема конфигурирования на алгоритмическом уровне потребовала создания модульной архитектуры и декларативного подхода к определению процесса извлечения. Модули получили название обрабатывающих ресурсов в противовес лингвистическим ресурсам и ресурсам знаний. В конфигурации декларируется порядок обработки документа аналитическими модулями, потоки данных между ними, а также параметры их работы.

Обрабатывающие ресурсы можно разделить на следующие группы.

- Ресурсы предобработки. Сюда относятся средства определения кодировки документа, извлечения текста и стилевой разметки из документа, предварительной фильтрации.
- Ресурсы лингвистического анализа. Осуществляют разбор текста на отдельные слова, морфологический анализ (в том числе специализированные варианты для различных категорий имен собственных), поверхностный синтаксический анализ и определение границ предложений.
- Ресурсы извлечения. Осуществляют поиск в документе целевой лексики и синтаксических конструкций, а также первичное структурирование информации.
- Ресурсы унификации знаний и вывода. Осуществляют унификацию и отождествление элементов знаний, вывод производных знаний.
- Ресурсы подготовки результата. Осуществляют приведение извлеченной информации к определенному формату и передачу за пределы последовательности обработки (в БД, глобальный ресурс знаний, файл, приложение).

В системе ИСИДА-Т все модули, в том числе средства общего лингвистического анализа, используют структуру данных – аннотация. Аннотация – объект, который приписывается фрагменту текста (например, слову, словосочетанию, предложению, ссылке на сущность предметной области и т. д.) и описывает свойства этого фрагмента. Аннотации разбиты на конечное множество классов. Каждый класс аннотаций описывает текст в определенном аспекте. Информация о фрагменте представлена значениями именованных атрибутов аннотации. Наборы классов и атрибутов аннотаций намеренно не специфицированы, чтобы можно было использовать произвольный набор обрабатывающих модулей и представлять необходимую лингвистическую и предметную информацию. Обмен данными между модулями тоже идет в терминах аннотаций: новые аннотации могут строиться на основании полученных на предыдущих этапах анализа [Овдей, 2006]. В реализации системы ИСИДА-Т модель аннотаций была дополнена некоторыми полезными средствами. В частности, было снято ограничение на атомарность атрибутов и добавлена возможность устанавливать ссылки между аннотациями.

Для распознавания текстовых ситуаций используется набор правил, описывающих характерные для конкретной задачи способы выражения ситуации в тексте. Эти правила задают образец для сопоставления и действия, которые должны быть произведены после успешного сопоставления. Ряд современных систем извлечения информации (в том числе, система ИСИДА-Т) берут за основу различные диалекты языка CPSL [Noy, 1999]. Использование этого языка подразумевает разметку текста при помощи аннотаций.

Язык правил, используемый в системе ИСИДА-Т, является расширением CPSL. Предлагаемые расширения преследуют две цели: 1) обеспечить возможность описывать более сложные контексты, в которых встречается целевая информация, и 2) снизить объем рутинной работы при создании системы правил за счет более компактного описания контекста [Гаврилова, 2000].

Отличия от других реализаций, например, JAPE [Calvanese, 2007] или диалекта CPSL состоят в следующем.

- Реализована встроенная поддержка расширенного спектра типов данных, в том числе, ссылок на аннотации и множественных значений. Данные этих типов могут использоваться в качестве значений переменных и значений атрибутов аннотаций.
- Логика работы интерпретатора правил приведена в максимальное соответствие поведению интерпретатора обычных регулярных выражений. Отличия от современной реализации JAPE и Montreal transducer [Calvanese, 2007] заключаются в поддержке «жадных» и «нежадных» квантификаторов и опережающей проверки.
- Поддерживаются кванторы существования (по умолчанию) и всеобщности, связывающие элементарные тесты. К кванторам может добавляться отрицание.
- Существуют языковые средства, позволяющие гибко проверять взаимное расположение аннотаций, рассматриваемых в контексте сопоставления, и прочих аннотаций во входной коллекции.
- В тестах могут использоваться функции для обращения к ресурсу знаний, например, проверки таксономической принадлежности элементов. Для более сложных запросов к ресурсу знаний используется предметно-ориентированный язык, совпадающий с языком описания левой части правил трансформации.
- Для передачи информации между элементарными тестами, а также в правую часть правил могут использоваться именованные переменные, значения которых присваиваются явно в ходе сопоставления. Множество значений переменных входит в контекст сопоставления.

## Инструментальное средство проектирования онтологий Protégé

Protégé [Noy, 1999] – локальная, свободно распространяемая Java-программа, разработанная группой медицинской информатики Стэндфордского университета. Программа предназначена для построения (создания, редактирования и просмотра) онтологий прикладной области. Её первоначальная цель – помочь разработчикам программного обеспечения в создании и поддержке явных моделей предметной области и включение этих моделей непосредственно в программный код. Protégé включает редактор онтологий, позволяющий проектировать онтологии, разворачивая иерархическую структуру абстрактных или конкретных классов и слотов. Структура онтологии сделана аналогично иерархической структуре каталога. На основе сформированной онтологии, Protégé может генерировать формы получения знаний для введения экземпляров классов и подклассов. Инструмент имеет графический интерфейс, удобный для использования неопытными пользователями, снабжен справками и примерами.

Protégé основан на фреймовой модели представления знания OKBC (Open Knowledge Base Connectivity) [Chaudhri, 1998] и снабжен рядом плагинов, что позволяет адаптировать его для редактирования моделей, хранимых в разных форматах (стандартный текстовый, в базе данных JDBC, UML, языков XML, XOL, SHOE, RDF и RDFS, DAML+OIL, OWL).

*Используемые формализмы и форматы*

Изначально единственной моделью знаний, поддерживаемой Protégé, была фреймовая модель. Этот формализм сейчас является «родным» для редактора, но не единственным.

Protégé имеет открытую, легко расширяемую архитектуру и, помимо фреймов, поддерживает все наиболее распространенные языки представления знаний (SHOE, XOL, DAML+OIL, RDF/RDFS, OWL). Protégé поддерживает модули расширения функциональности (plug-in). Расширять Protégé для использования нового языка проще, чем создавать редактор этого языка «с нуля».

Protégé основан на модели представления знаний OKBC (Open Knowledge Base Connectivity). Основными элементами являются классы, экземпляры, слоты (представляющие свойства классов и экземпляров) и фасеты (задающие дополнительную информацию о слотах).

*Пользовательский интерфейс*

Пользовательский интерфейс состоит из главного меню и нескольких вкладок для редактирования различных частей базы знаний и ее структуры. Набор и названия вкладок зависят от типа проекта (языка представления) и могут быть настроены вручную. Обычно имеются следующие основные вкладки: Классы, Слоты (или Свойства для OWL), Экземпляры, Метаданные.

## Инструментальный комплекс автоматизированного построения онтологий ПрО

Инструментальный комплекс онтологического назначения (ИКОН) для автоматизированного построения онтологии в произвольной предметной области [Палагин, 2012] является системой, реализующей одно из направлений комплексных технологий Data & Text Mining, а именно – анализ и обработку больших объёмов неструктурированных данных, в частности лингвистических корпусов текстов на украинском и/или русском языке, извлечение из них предметных знаний с последующим их представлением в виде системно-онтологической структуры или онтологии предметной области.

Извлечение информации (Information Extraction) [Палагин, 2012] — это подход, позволяющий сузить круг задач, требующих специфического предметно-ориентированного решения при анализе текста. В рамках этого подхода задача обработки текста ограничена распознаванием множества классов ключевых понятий конкретной предметной области и игнорированием всякой другой информации.

Несмотря на то, что системы извлечения информации могут строиться для выполнения различных задач, подчас сильно отличающихся друг от друга, существуют компоненты, которые можно выделить практически в каждой системе.

В состав почти каждой системы извлечения информации входят четыре основных компонента, а именно: компонент разбиения на лексемы, некоторый тип лексического или морфологического анализа, синтаксический анализ (микро- и макроуровень), модуль извлечения информации и модуль для анализа на уровне конкретной предметной области. В зависимости от требований к конкретному программному продукту в приведённую выше схему добавляют дополнительные модули анализа (специальная обработка составных слов; устранение омонимии; выделение составных типов, которое может также быть реализовано на языке правил извлечения информации; объединение частичных результатов).

Разбиение на слова при анализе европейских языков не является проблемой, поскольку слова отделяются друг от друга пробелом (или знаками препинания). Тем не менее, для обработки составных слов, аббревиатур, буквенно-цифровых комплексов и ряда других особых случаев требуются специфические алгоритмы. С границами предложений, как правило, тоже больших проблем не возникает. Однако при анализе таких языков как японский или китайский, определение границ слова на основе орфографии невозможно. По этой причине системы извлечения информации, работающие с такими языками, должны быть дополнены модулем сегментирования текста на слова.

В некоторые системы наряду с обычными средствами лексического и морфологического анализа могут быть включены модули для определения и категоризации атрибутов частей речи, смысловых нагрузок слов, имен или других нетривиальных лексических единиц.

ИКОН предназначен для реализации множества компонентов единой информационной технологии:

- поиск в сети Интернет и/или в других электронных коллекциях (ЭлК) текстовых документов (ТД), релевантных заданной ПрО, их индексацию и сохранение в базе данных;
- автоматическая обработка естественно-языковых текстов;
- извлечение из множества ТД знаний, релевантных заданной ПрО, их системно-онтологическая структуризация и формально-логическое представление на одном (или нескольких) из общепринятых языков описания онтологий. Кроме того, внутри этой технологии реализуется процедура построения, визуализации и проверки семантических структур синтаксических единиц ТД и понятийных структур заданной ПрО в виде онтографа, названного начальной онтологией ПрО;
- создание, накопление и использование больших структур онтологических знаний в соответствующих библиотеках;
- системная интеграция онтологических знаний как одна из основных компонент методологии междисциплинарных научных исследований;
- другие процедуры, связанные с автоматизацией приобретения знаний из множества естественно-языковых объектов.

ИКОН состоит из трёх подсистем и представляет собой интеграцию разного рода информационных ресурсов (ИР), программно-аппаратных средств обработки и процедур естественного интеллекта (ЕИ), которые, взаимодействуя между собой, реализуют совокупность алгоритмов автоматизированного, итерационного построения понятийных структур предметных знаний, их накопления и/или системной интеграции. Обобщённая блок-схема ИКОН представлена на рис. 2.

Подсистема «Информационный ресурс» включает блоки формирования лингвистического корпуса текстов, баз данных языковых структур и библиотек понятийных структур. Первый компонент представляет собой различные источники текстовой информации, поступающей на обработку в систему. Второй компонент представляет собой различные базы данных обработки языковых структур, часть из которых формируется (наполняется данными) в процессе обработки ТД, а другая часть формируется до процесса построения онтологии ПрО и, по сути, является ЭлК различных словарей. Третий компонент представляет собой совокупность библиотек понятийных структур разного уровня представления (от наборов терминов и понятий до высокоинтегрированной онтологической структуры междисциплинарных знаний) и является результатом реализации некоторого проекта (проектирования онтологии ПрО и/или системной интеграции онтологий).

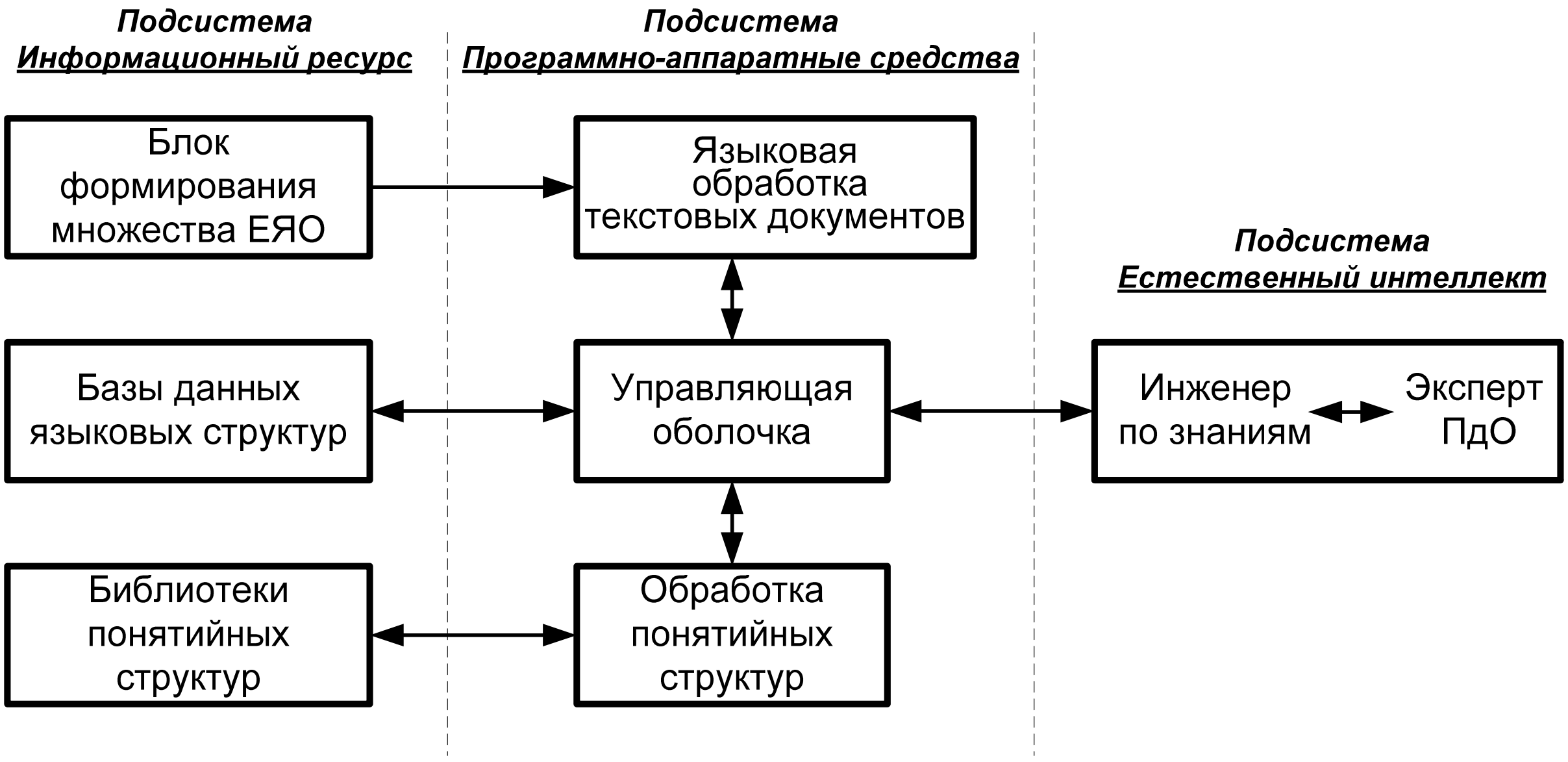

Рис. 2. Обобщённая блок-схема ИКОН

Подсистема «Программно-аппаратные средства» включает блоки обработки языковых и понятийных структур и управляющую графическую оболочку, с помощью которой инженер по знаниям осуществляет общее управление процессом использования связанных информационных технологий.

Подсистема «Естественный интеллект» осуществляет подготовку и реализацию процедур предварительного этапа проектирования, а на протяжении всего процесса осуществляет контроль и проверку результатов выполнения этапов проектирования, принимает решение о степени их завершённости (и в случае необходимости – повторении некоторых из них).

## ТОДОС – IT-платформа онтологических информационно-аналитических экспертных систем

ТОДОС – инновационный комплекс программно-информационных и методических средств управления знаниями с использованием подходов онтологического управления корпоративными информационными ресурсами, где человек рассматривается как источник определения новых знаний для передачи их в форме собственного знания через инструментарий [Стрижак, 2013]. ТОДОС обеспечивает для пользователя единую интегрированную точку доступа – «единое окно» – к информации и приложениям системы для обеспечения интерактивного взаимодействия при решении прикладных задач.

Принцип ТОДОС – «Ситуационная осведомленность» за счет предоставления пользователям необходимой информации, касающейся направлений их деятельности и достаточной для принятия эффективного решения:

- Удобное, интуитивно-понятное, многоаспектное представление аналитической информации;
- Обеспечение работы с неструктурированной и слабо структурируемой информацией;
- Работа с информацией и результатами анализа из любой точки сетевого доступа;
- Удовлетворение поиска и запросов пользователей – извлечение знаний;
- Обработка и анализ контента, агрегирование и упорядочивание по заданному критерию;
- Поддержка принятия решений на основе анализа больших объемов информации;
- От данных - к пространственно-распределённым системам управления (ГИС) - от ГИС - к информации;
- Обеспечение взаимодействия и обратной связи.

Концепция ТОДОС:

- консолидация и интеграция всей имеющейся корпоративной информации и предоставление ее через систему «единого окна», за счет чего повышается уровень осведомленности всех категорий пользователей в их деятельности;
- обеспечение «бесшовной» системной интеграции информационных технологий и инноваций с целью создания информационно-аналитических ресурсов для внедрения в бизнес-процессы организации;
- создание условий «ситуационной осведомленности» для всех заинтересованных категорий пользователей с многоаспектным анализом массивов документов, их анализом, сравнением, рейтингованием с выводом отчетов и результатов анализа;
- обеспечение онтологического управления информационными массивами, которые объединяются в единое корпоративное информационное пространство – онтолого-управляемую систему корпоративных знаний;
- поиск в сети Интернет и в файловых электронных коллекциях текстовых документов, релевантных тематике исследований и экспертизы;

- автоматическая обработка естественно-языковых текстов с выделением поверхностных семантических отношений для дальнейшего их анализа;
- извлечение из множества документов знаний, релевантных выбранной предметной области, их системно-онтологическая структуризация и формально-логическое представление, а также построение, визуализация и верификация семантических структур синтаксических единиц текстовых документов и категориальных знаний заданной предметной области в виде онтологического графа;
- автоматизированное построение онтологий и тезаурусов предметных областей для организации системы управления знаниями;
- автоматизированный анализ и создание системы рейтингов объектов исследования и процессов с ними связанных с учетом всего множества факторов, влияющих на соответствующие объекты и процессы;
- обеспечение многовекторного исследования объектов и процессов с целью выявления параметров влияющих на их состояние, развитие и принятия соответствующего объективного решения.

ТОДОС – это:

- Модуль КОНСПЕКТ – контекстно-семантический анализ естественно-языкового текста и построение таксономии документов;
- Модуль КОНФОР – классификация и генерация онтологий предметной области;
- Модуль ЭДИТОР – конструирование трансдисциплинарных онтологий;
- Модуль АЛЬТЕРНАТИВА – онтология задачи выбора для информационно-аналитической поддержки принятия решений и обеспечения процессов многофакторного анализа и рейтингования;
- ПОИСКОВАЯ МАШИНА – поиск лексических структур на основе лингвистической обработки большого количества распределенных сетевых текстовых массивов;
- ЛИНГВИСТИЧЕСКИЙ КОРПУС – электронная библиотека со средствами ассоциативного поиска семантически связанных информационных массивов;
- Модуль КРИПТО – защита информационных массивов от несанкционированного доступа.

Модуль КОНСПЕКТ представляет собой лингвистический процессор [Величко, 2009], который обеспечивает первичное формирование лингвистического корпуса [Широков, 2005] и позволяет решать следующие практические задачи:

1) Выделение однословных и многословных терминов документа.

2) Формирование списка терминов с подсчетом их количества в тексте.

3) Навигация по списку терминов с отбором и отражением контекстов предложений из текста с выделенными терминами.

4) Автоматизированное построение тематического словаря по коллекции документов для использования в локальной полнотекстовой поисковой системе с целью повышения релевантности поиска.

5) Конспектирование текста по заданной теме. Конспект – это краткое изложение содержания статьи или книги. Система автоматически создает текстуальный конспект, где основные положения произведения, доказательства и выводы передаются словами автора, то есть цитатами. Тема конспекта задается пользователем программы с помощью ключевого слова или словосочетания. Из исходного текста в конспект отбираются предложения, содержащие тему, и автоматически отбираются термины текста, ассоциативно связанные с заданной темой. Пример конспекта, который автоматически построен в среде модуля КОНСПЕКТ, приведен на рис. 3.

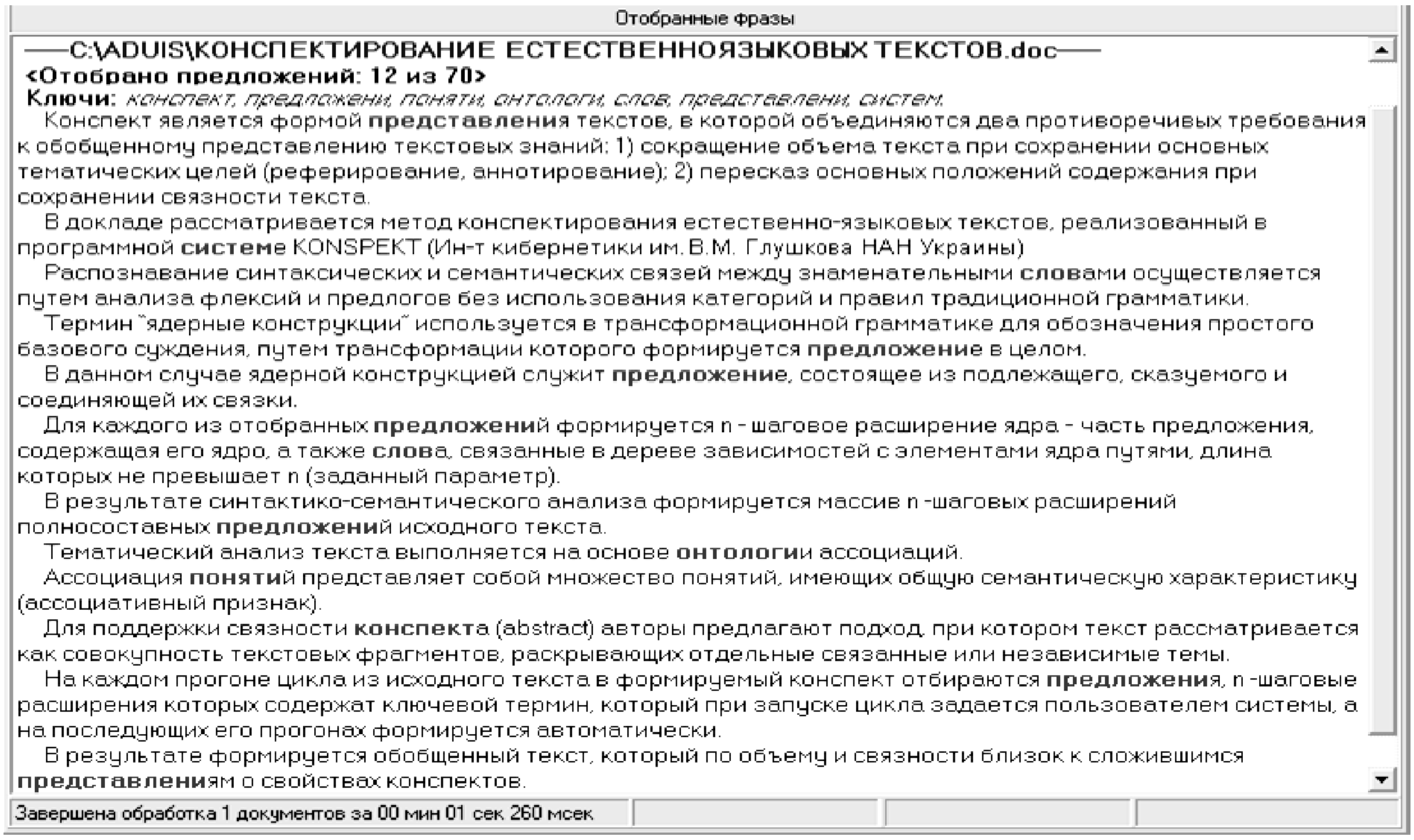

Рис. 3. Фрагмент среды модуля КОНСПЕКТ

Результаты, которые можно получить в ближайшее время при условии доработки разработанной программы:

1) Повышение качества обработки текстов на украинском и русском языках за счет увеличения объема словаря системы;

2) Автоматическое определение тематических направлений документа;

3) Сортировка документов по тематическим направлениям;

4) Построение семантической сети терминов документа;

5) Объединение семантических сетей терминов для нескольких документов;

6) Поиск информации в сети Интернет с помощью поискового модуля системы ТОДОС с дополнительной автоматической семантической фильтрацией результатов поиска. Для дополнительной фильтрации используются алгоритмы формирования семантических категорий заданной тематики, извлеченных из текста. Список документов, которые получает пользователь после дополнительной фильтрации, сокращается в несколько раз;

7) Конспектирование текста по заданной теме (для украинского языка).

Модуль КОНФОР предназначен для построения семантических связей между объектами предметной области на основе индуцированной обработки отношений между объектами тематических классов ПрО. Для этого тематические классы описываются пользователем в соответствии с нижеприведённой схемой:

(имя класса объектов 1),(имя связи), (имя объекта 1),...,(имя объекта n)

(имя класса объектов 2),(имя связи), (имя объекта 1),...,(имя объекта j)

.........................................................................................................................

(имя класса объектов m),(имя связи), (имя объекта 1),...,(имя объекта k)

Согласно схемы заполняется таблица, например MS Excel, ячейки столбца A которой содержат имена материнских вершин графа, ячейки столбца B – имена связей между вершинами, ячейки столбцов от C до Z – имена дочерних вершин. Выходной структурой модуля КОНФОР является онтограф ПрО [Палагин, 2012], [Довгий, 2013], вершины которого носят имена объектов.

Модуль ЭДИТОР предназначен для создания, редактирования, просмотра и анализа сетей понятий и формирования закономерностей, представленных в виде набора значений признаков, которыми описываются начальные понятия предметной области. Таким образом, описание предметной области является наглядным и легко интерпретируется пользователем.

Выделение закономерностей происходит методом индуктивного формирования понятий на основе пирамидальной сети.

Пирамидальной сетью называется ациклический ориентированный граф, вершинами которого выступают термины (понятия) предметной области, а ребрами (дугами) – семантические связи между терминами [Гладун, 1994].

Представление информации в виде графа позволяет изучать не только отдельный термин (понятие), но и получать все его семантические связи с другими понятиями, тем самым осмысливая его роль в данной системе знаний или в ходе решения задачи.

Однако сетевой граф может выступать не только средством организации знаний. Расширяя традиционные функции, граф можно превратить в среду, в которой обеспечивается активная работа со знаниями, а также оригинальным образом решаются учебные задачи.

Концептуальная модель предметной области или онтология состоит из графа (иерархии понятий) предметной области, связей между ними и законов, которые действуют в рамках этой модели.

Модуль АЛЬТЕРНАТИВА обеспечивает упорядочивание объектов-концептов онтологии, на основе интегрированной обработки характеризующих их свойств. Для этого используются весовая, бальная и лингвистическая шкалы. Каждая такая шкала определяет значение критериев, характеризующих свойства объектов тематической онтологии ПрО. В общем случае свойства-критерии характеризуются различными степенями важности, которые при решении задачи выбора задаются некоторыми действительными числами – весовыми коэффициентами. Перед решением задачи для каждого критерия необходимо сформировать его значение для каждой альтернативы [Саати, 1989].

Модули ПОИСКОВАЯ МАШИНА и ЛИНГВИСТИЧЕСКИЙ КОРПУС, обеспечивают маркировку и индексирование семантических единиц, определяющих и описывающих контексты объектов тематических онтологий ПрО. Контексты семантических единиц составляют электронную библиотеку.

Модуль КРИПТО обеспечивает защиту информационных ресурсов электронной библиотеки системы. Защита реализуется на основе гомоморфного кодирования. Для создания эффективных методов защиты системы реализуются механизмы инфраструктуры публичных ключей (PKI). Инфраструктура публичных ключей позволяет объединить электронные подписи, публичные и частные ключи и пирамиду виртуальных лиц или организаций, которые могут выполнять действия по управлению тематическими онтологиями и отображающими их документами.

На основе анализа сопроводительной документации, справочных материалов, доступных в открытых источниках информации (интернет, научные труды), и, в отдельных случаях, практической работы с программными системами конструирования онтологий (ИКОН, ТОДОС, Protégé), построена сводная таблица, обобщающая сравнительные технические характеристики некоторых специализированных инструментальных средств инженерии онтологий (таблица 1).

## Заключение

В статье представлен обзор специализированных инструментальных средств инженерии онтологий. Как видно из сравнительной таблицы, в которой представлены различные системы конструирования онтологий, стратегическим направлением развития таких систем является интероперабельность. Эффективность использования онтологических описаний в основном зависит от возможностей применения форматов их описаний при решении различных сложных прикладных задач. Наиболее конструктивным методом решения этой проблемы может быть разработка онтологий задач, которые смогут обеспечить семантическую синхронизацию взаимодействий различных по тематикам информационных систем на основе установления отношений между свойствами их концептов.

## Библиография

**Таблица 1 Системы конструирования онтологий**

| | ИКОН | ЛоТА | SIMER+MIR | ИСИДА-Т | Protégé | Ontolingua | InTez | OntoSTUDIO | OntoEdit | ТОДОС |
|---|---|---|---|---|---|---|---|---|---|---|
| **Общая информация** | | | | | | | | | | |
| Разработчик | ИК НАНУ | ? | РАН ИПС ДЦШI | РАН ИПС ДЦШI | SMI, Stanford University | KSL, Stanford University | С-ПТ ДУ | Semafora systems GmbH | Ontoprizse | НЦ МАНУ, ИК НАНУ, ИТГИП НАНУ, НТУ КПИ, НаУКМА |
| Версия | 1.0 | ? | ? | ? | 4.1 | 4.0 | ? | 3.1 | | 3.0 |
| URL | http://aduis.com.ua/ | – | – | http://isida-t.ru/ | http://protege.stanford.edu/ | http://www.ksl.stanford.edu/software/ontolingua/ | http://inttez.ru | http://www.semafora-systems.com/en/products/ontostudio | http://www.semafora-systems.com/products/ontoedit | http://editor3/inhost.com.ua |
| Доступность | Закрытый код, ограниченный доступ | Закрытый доступ | Закрытый доступ | Закрытый доступ | Закрытый код, свободный доступ | Свободный доступ | Закрытый код, свободный доступ | Закрытый код, ограниченный доступ | Свободная лицензия | Открытый код, свободный доступ с разделением прав |
| Интерфейс пользователя | Локальное приложение | ? | Локальное приложение | Локальное приложение | Локальное приложение, Веб-приложение | Веб- приложение | Локальное приложение | Локальное приложение, Веб-приложение | Локальное приложение | Веб-приложение |
| Язык интерфейса ПО | Русский, Украинский | Русский | Русский | Русский, Английский | Английский | Английский | Русский | Английский | Английский | Украинский, Русский, Английский, Арабский |
| **Архитектура программного обеспечения** | | | | | | | | | | |
| Расширение | Плагины | – | – | – | Плагины | Плагины | – | Плагины | Плагины | Плагины |
| Язык ПО (платформа) | Java, Delphi | C++ | ? | ? | Java | Lisp | ? | Построенный на платформе Eclipse | Java | Java, Delphi, C++, HTML 5 |
| Сохранение онтологий | Файлы, СУБД | СУБД (формат TPS) | Файлы | Файлы | Файлы, СУБД | Файлы | Файлы, СУБД | Файлы, СУБД | Файлы | Файлы, СУБД |
| Архитектура приложения | n-уровневая | Клиент-сервер | | 3-х уровневая | 3-х уровневая | Клиент-сервер | | 3-х уровневая | 3-х уровневая | Клиент-сервер |
| **Функционал инструментального средства** | | | | | | | | | | |
| Язык обрабатываемых ЛКТ | Русский, Украинский | Русский | Русский | Русский | Английский | Английский | Русский | – | – | Украинский, Русский, Английский |

| | ИКОН | ЛоТА | SIMER+MIR | ИСИДА-Т | Protégé | Ontolingua | InTez | OntoSTUDIO | OntoEdit | ТОДОС |
|---|---|---|---|---|---|---|---|---|---|---|
| Управляющая графическая оболочка системы | + | + | + | + | + | + | + | + | + | + |
| Системная интеграция онтологий | + | – | | – | + | – | – | + | + | + |
| Система поиска ТД (Поиск ТД во внешних и внутренних информационных ресурсах) | + | – | | – | – | + | | | – | + |
| Визуальное проектирование, редактирование онтологических структур | + | + | | + | + | + | | + | + | + |
| Автоматизированное построение онтологий ПдО | +– | – | | – | - | – | | – | – | + |
| Ручное построение онтологии ПдО | + | – | | + | + | + | | + | + | + |
| Автоматический лингвистический анализ ТД, описывающих заданную ПдО (синтактико-семантический, морфосинтаксический анализ) | + | + | | + | - | – | | – | – | + |

| | ИКОН | ЛоТА | SIMER+MIR | ИСИДА-Т | Protégé | Ontolingua | InTez | OntoSTUDIO | OntoEdit | ТОДОС |
|---|---|---|---|---|---|---|---|---|---|---|
| Формально-логическое представление и интеграция онтологических структур в онтологическую базу знаний ПдО | + | + | | + | + | + | | + | + | + |
| Сохранение и управление электронными коллекциями энциклопедических и толковых словарей, тезаурусов | + | + | | + | - | + | | – | – | + |
| Лингвистический корпус текстов (ЛКТ) (База данных ТД) | + | + | | + | - | – | | + | – | + |
| Объектное отображение таксономий | – | – | – | – | – | – | – | – | – | + |
| Библиотека онтологий задач | – | – | – | – | – | – | – | – | – | + |
| Эксперт ПдО | + | + | + | + | + | + | + | + | + | + |
| Инженер по знаниям | + | + | + | + | + | + | + | + | + | + |
| Поддержка взаимодействия экспертов | – | – | – | – | – | – | – | – | – | + |

| | ИКОН | ЛоТА | SIMER+MIR | ИСИДА-Т | Protégé | Ontolingua | InTez | OntoSTUDIO | OntoEdit | ТОДОС |
|---|---|---|---|---|---|---|---|---|---|---|
| **Интероперабельность** | | | | | | | | | | |
| Конвертация файлов описания онтологий | + | – | – | – | + | KIF, Prolog, Ontolingua, LOOM, CLIPS, IDL | Экспорт в OWL | Импорт из UML 2.0, Database schemas (Oracle, MS-SQL, DB2, MySQL), Excel tables, Outlook E-Mails, Folder structures of the file system | Импорт из OXML, RDF(S), DAML+OIL, Flogic Экспорт в OXML, RDF(S) | Импорт из OWL, OXML, RDF(S), FLogic, Excel tables, Экспорт в OWL, OXML, RDF(S), Excel tables, Prolog, CLIPS, IDL |
| Интеграция с другими инструментальным и средствами | Protégé | ? | ? | ? | Возможна средствами плагинов | OKBC, Chimaera | | Protégé | OntoAnnotat, OntoMat, Semantic-Miner | Protégé - возможна средствами плагинов |
| Интеграция с базой знаний WolframAlpha | + | – | – | – | – | – | – | – | – | + |
| Интеграция с поисковыми системами : Google, Bing, Exalead, Convera, Галактика Zoom, Информбюро и т.д. | - | – | – | – | – | – | – | – | – | + |
| Интеграция с пространственно-распределёнными системами управления (ГИС-системы) | – | – | – | – | – | – | – | – | – | + |
| Параллельная обработка данных | + | – | – | + | – | – | – | – | – | + |

## Информация об авторах

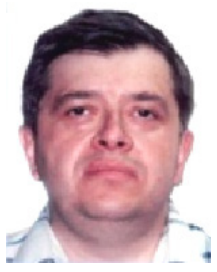

***Величко Виталий Юрьевич*** *– к.т.н., старший научный сотрудник, Институт кибернетики им. В. М. Глушкова НАН Украины, Киев-187 ГСП, 03680, просп. акад. Глушкова, 40; e-mail: velychko@aduis.com.ua*

*Основные области научных исследований: индуктивный логический вывод, обработка естественно-языковых текстов*

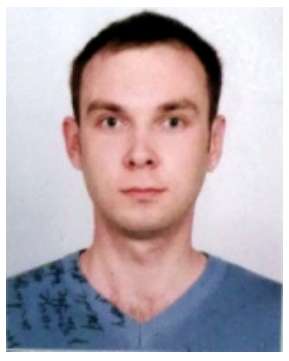

***Малахов Кирилл Сергеевич*** *– младший научный сотрудник, Институт кибернетики им. В. М. Глушкова НАН Украины, Киев-187 ГСП, 03680, просп. акад. Глушкова, 40; e-mail: malahov@live.com*

*Основные области научных исследований: онтологический инжиниринг*

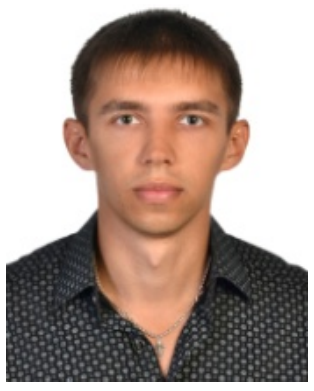

***Семенков Виталий Васильевич*** *– аспирант, Луганський национальний университет имени Тараса Шевченка*

*Основные области научных исследований: онтологический инжиниринг*

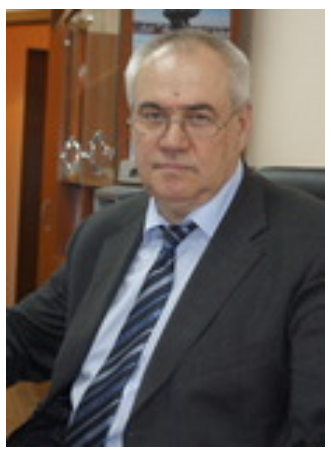

***Стрижак Александр Евгеньевич*** *– к.т.н., старший научный сотрудник, Институт телекоммуникаций и глобального информационного пространства НАН Украины, Киев-186, 03186, Чоколовский бульвар, 13; e-mail: sae953@gmail.com*

*Основные области научных исследований: корпоративные интеллектуальные системы, поддержка принятия решений*

## Integrated Tools for Engineering Ontologies

**Velychko V.Yu., Malahov K.S., Semenkov V.V., Strizhak A.E.**

***Abstract****: The article presents an overview of current specialized ontology engineering tools, as well as texts' annotation tools based on ontologies. The main functions and features of these tools, their advantages and disadvantages are discussed. A systematic comparative analysis of means for engineering ontologies is presented.*